\documentclass{article} 
\usepackage{iclr2025_conference,times}

\usepackage{amsmath,amsfonts,bm}

\def\eqref#1{equation~\ref{#1}}

\def\1{\bm{1}}

\DeclareMathAlphabet{\mathsfit}{\encodingdefault}{\sfdefault}{m}{sl}
\SetMathAlphabet{\mathsfit}{bold}{\encodingdefault}{\sfdefault}{bx}{n}

\usepackage{hyperref}
\usepackage{url}
\usepackage{tabularx}
\usepackage{booktabs}
\usepackage{caption}
\usepackage{adjustbox}
\usepackage{makecell}
\usepackage{enumitem}
\usepackage{graphicx}
\usepackage{xspace}
\usepackage{fancyhdr}
\usepackage{tikz}
\usepackage{float}
\usetikzlibrary{shapes.geometric, arrows.meta, positioning, decorations.pathreplacing}
\usepackage[T1]{fontenc}

\newcommand{\ourmodel}{JT-Math-8B\xspace}

\title{ \jiutian   JT-Math: A Multi-Stage Framework for Advanced Mathematical Reasoning in Large Language Models}

\author{
Yifan Hao\footnotemark[1], Fangning Chao\footnotemark[1], Yaqian Hao\footnotemark[1], Zhaojun Cui\footnotemark[1], Huan Bai\footnotemark[1], Haiyu Zhang\footnotemark[1],\\
\textbf{  Yankai Liu} \textsuperscript{\footnotemark[1]}          \textsuperscript{\footnotemark[2]} \textbf{, Chao Deng ,} and \textbf{Junlan Feng}\footnotemark[2] \\
\textit{JIUTIAN Team, China Mobile Research Institute}\\
Beijing, China  \\
\texttt{\{liuyankai, fengjunlan\}@chinamobile.com}
}

\newcommand{\jiutian}{\raisebox{-2.5 pt}{\includegraphics[height=1.2em]{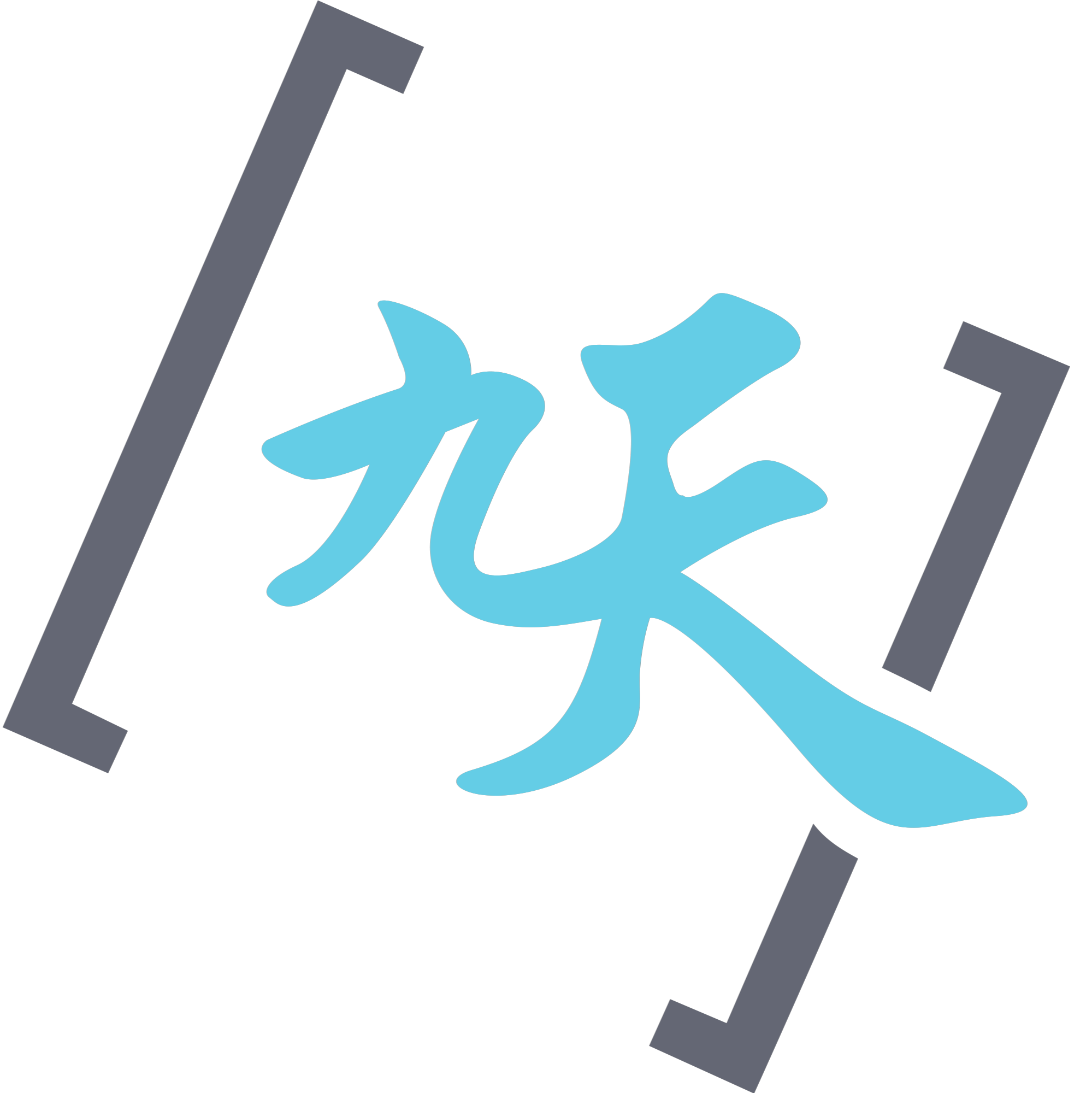}}\xspace}

\fancypagestyle{firstpageicon}{
    \fancyhf{} 
    \fancyhead[L]{\includegraphics[height=2em]{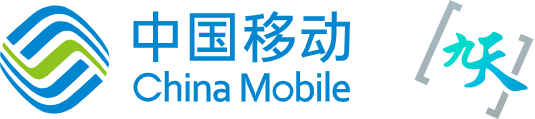}} 
}

\iclrfinalcopy 
\begin{document}
\thispagestyle{firstpageicon}

\maketitle
\footnotetext[1]{Equal Contribution.}
\footnotetext[2]{Co-corresponding authors.}

\begin{abstract}
Mathematical reasoning is a cornerstone of artificial general intelligence and a primary benchmark for evaluating the capabilities of Large Language Models (LLMs). While state-of-the-art models show promise, they often falter when faced with complex problems that demand deep conceptual understanding and intricate, multi-step deliberation. To address this challenge, we introduce \ourmodel, a series of open-source models comprising base, instruct, and thinking versions, built upon a systematic, multi-stage optimization framework. Our pre-training corpus is a high-quality, 210B-token dataset curated through a dedicated data pipeline that uses model-based validation to ensure quality and diversity. The Instruct Model is optimized for direct, concise answers through Supervised Fine-Tuning (SFT) and a GRPO-based reinforcement learning (RL) method. The Thinking Model is trained for complex problem-solving using a Long Chain-of-Thought (Long CoT) approach, combining SFT with a novel, multi-stage RL curriculum that progressively increases task difficulty and context length up to 32K tokens. \ourmodel achieves state-of-the-art results among open-source models of similar size, surpassing prominent models like OpenAI's O1-mini and GPT-4o , and demonstrating superior performance on competition-level mathematics.
\end{abstract}

\enlargethispage{2.5\baselineskip}
\begin{figure}[htbp]
    \vspace{-5pt}
    \centering
    \includegraphics[width=\linewidth]{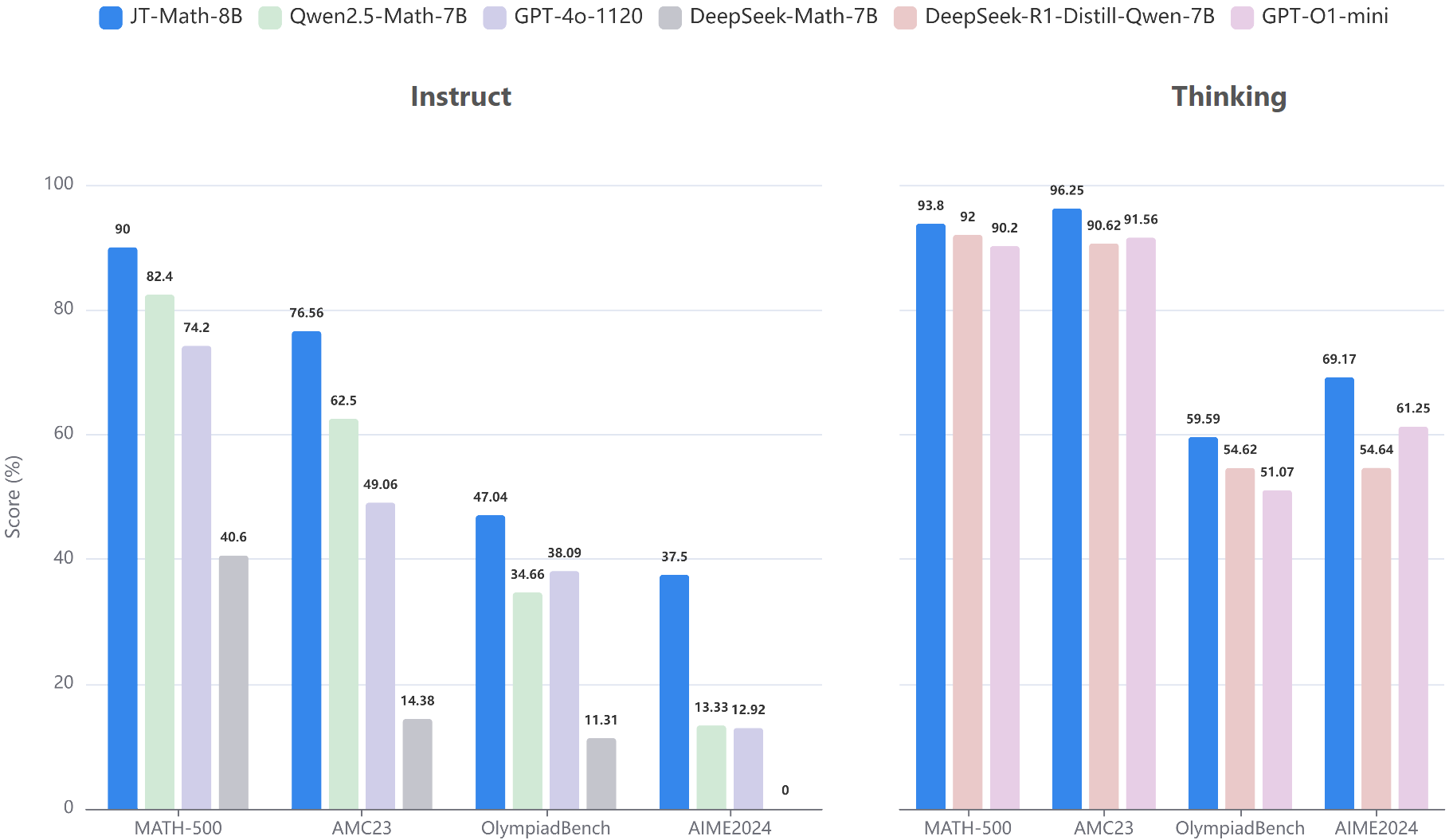}
    \label{fig:teaser}
    \vspace{-17pt}
    \caption{Benchmark performance of instruct and thinking variants of \ourmodel.}
\end{figure}

\newpage

\pagestyle{fancy}
\fancyhf{}
\fancyfoot[C]{\small \thepage}
\renewcommand{\headrulewidth}{0pt}

\tableofcontents

\newpage

\section{INTRODUCTION}

Mathematical reasoning, which encompasses logical deduction, abstract generalization, and multi-step problem-solving, is widely regarded as a cornerstone for the development of Artificial General Intelligence (AGI). Therefore, it has become a key benchmark for testing the cognitive abilities of Large Language Models (LLMs). While state-of-the-art LLMs have demonstrated considerable promise in solving mathematical problems, underscoring the viability of this paradigm, their limitations become apparent when confronted with problems of greater difficulty that demand deep conceptual understanding and intricate, multi-step deliberation. This gap highlights an urgent need to enhance the mathematical proficiency of these models. Therefore, developing methodologies to systematically cultivate and advance the deep mathematical reasoning of LLMs has emerged as a pivotal and formidable challenge in AI research.

In this context, the open-source community has produced a number of powerful models, such as Qwen 2.5 Math \citep{yang2024qwen25mathtechnicalreportmathematical}, DeepSeek Math \citep{shao2024deepseekmathpushinglimitsmathematical}, and DeepSeek-R1 \citep{deepseekai2025deepseekr1incentivizingreasoningcapability} , which provide robust baselines for mathematical capabilities. Concurrently, proprietary models, including OpenAI's O-series and Google's Gemini 2.5 Pro, continue to establish new performance benchmarks, demonstrating that proficiency in complex mathematics is a key differentiator. A closer examination reveals that most mathematics-focused models predominantly adhere to a shared training paradigm: initial pre-training on extensive corpora of mathematical and scientific text, succeeded by post-training phases such as Supervised Fine-Tuning (SFT) and Reinforcement Learning (RL), for instance, through methods like GRPO \citep{shao2024deepseekmathpushinglimitsmathematical}.

In this work, we propose a multi-stage framework for both pre-training and post-training, engineered to systematically build a model's mathematical prowess. We introduce \ourmodel, a series of large language models trained from scratch and purpose-built for mathematical tasks. During the pre-training phase, we employ a multi-faceted approach—including continued pre-training, an annealing stage, and context length extension—to specifically bolster the model's domain knowledge and processing capacity. In the post-training phase, we initiate with SFT and then apply a curriculum reinforcement learning method derived from GRPO to further refine its problem-solving abilities. Our approach yields models with mathematical skills that surpass those of prominent models like OpenAI's O1-mini and GPT-4o. We release our models in two primary versions: an Instruct model optimized for direct-answer generation, and a Thinking model that incorporates a "deep think" mechanism to deconstruct and solve more challenging problems.

\noindent Our core contributions are:

\begin{itemize} [leftmargin=*, topsep=0pt, itemsep=3pt]
    \item\textbf{Pre-Training:} Our research provides the development of a dedicated data processing pipeline for mathematical data, ensuring high-quality pretraining corpus. Through rigorous synthesis, filtering, and validation, we constructed a 210B tokens JT-Math Corpus with optimized diversity and difficulty distribution. Building upon this foundation, we conducted three-stage continued pretraining, progressively adapting the model to mathematical reasoning.  
    \item\textbf{Post-Training:} We architect a dual-path post-training strategy to produce two specialized models: an Instruct model for short-chain reasoning and a Thinking model for complex, long-chain thought. The process begins with the construction of two distinct Supervised Fine-Tuning (SFT) datasets, which are meticulously curated through a rigorous filtering pipeline to teach either short or long Chain-of-Thought (CoT). Following this foundational alignment, we employ a curriculum Reinforcement Learning (RL) in which tasks progressively increase in both complexity and required reasoning length, systematically cultivating advanced problem-solving skills while ensuring stable policy updates.
    
    \item\textbf{Open-Source:}The \ourmodel achieves state-of-the-art (SOTA) performance across multiple mathematical benchmark. We are releasing the entire JT-Math series to the public, including the Base, Instruct, and Thinking models. We believe that this report and our open-source models will provide useful insights to the mathematical LLM community and help encourage further development in this area.
\end{itemize}

\section{Pretraining}\label{sec:pretrain}
\subsection{Pre-training Data}
\begin{figure}[h]
    \centering
    \includegraphics[width=1.0\textwidth]{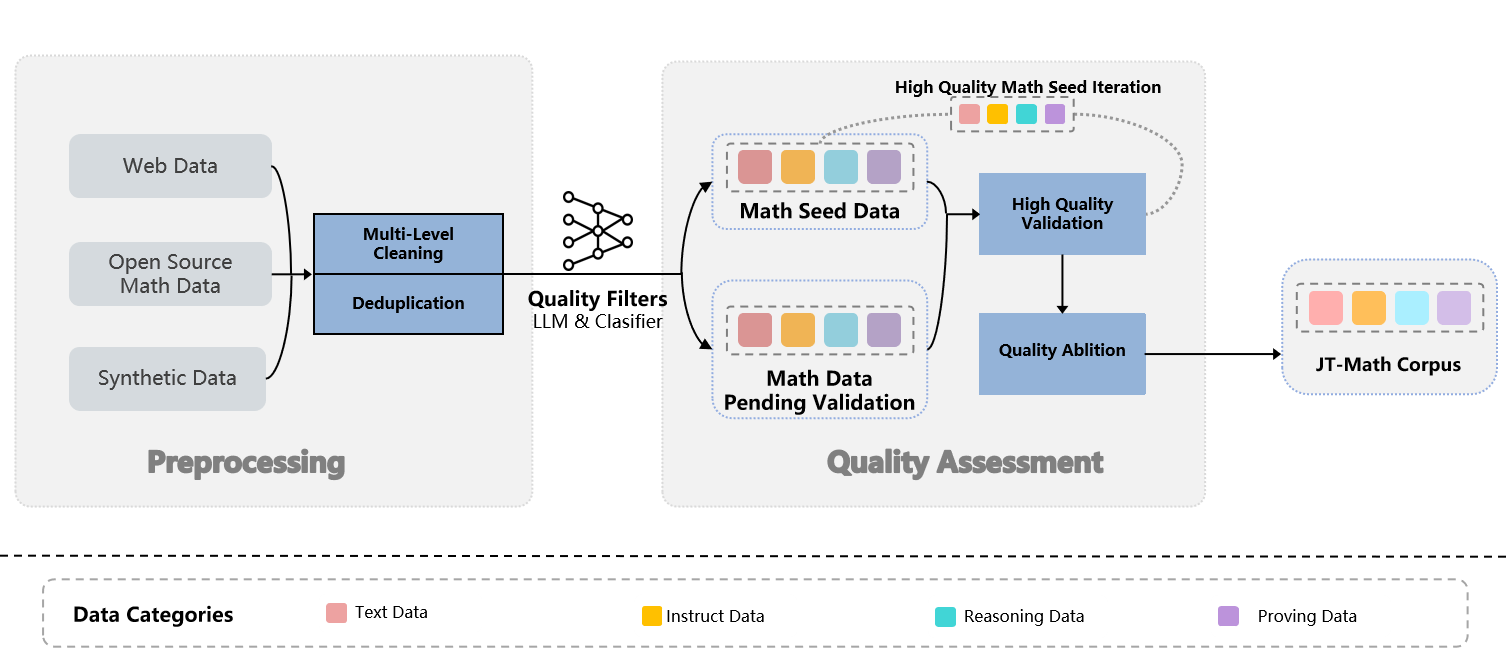}
    \caption{Data Processing Pipeline for JT-Math Corpus Construction}
    \label{fig:data_pipeline}
\end{figure}
\subsubsection{Data Pipeline}
In this section, we outline the data processing pipeline for JT-Math Corpus. As shown in Figure ~\ref{fig:data_pipeline}, our dataset integrates various sources, including web-crawled mathematical data, open source mathematical domain data (e.g., exam questions, encyclopedias, reasoning, theorem-proving), and LLM \citep{yang2024qwen25mathtechnicalreportmathematical} generated translations and synthetic data. Through systematic data collection and enhancement, we first apply cleaning, de-duplication, and quality filtering to produce candidate datasets. These then undergo iterative quality verification to ultimately construct our high-quality JT-Math Corpus, with verified accuracy, pedagogical rigor, linguistic fluency, and content diversity.

\paragraph{Data Collection and Enhancement}

To construct a comprehensive mathematical dataset, we systematically gather data from multiple high-quality sources, including rigorously screened public web resources and authoritative large-scale open-source mathematical datasets \citep{allal2025smollm2smolgoesbig, zhou2025megamath, tian2025deepdistillenhancingllmreasoning}. These datasets cover a broad spectrum of mathematical fields, ranging from fundamental to advanced levels. Given that most open source data is primarily in English while high quality Chinese mathematical data remains equally critical, we translate premium English datasets to generate an initial Chinese mathematical dataset.
\paragraph{Data Cleaning and Deduplication}
Recognizing that raw datasets often contain noise and problematic content, we implement a rigorous multi-stage cleaning pipeline. This includes length-based intelligent filtering to remove invalid samples, specialized text processing, and comprehensive standardization to ensure encoding consistency. For the translated Chinese mathematical data, all outputs undergo a three-step refinement process: rule-based secondary filtering, cleaning, and LLM-based \citep{yang2024qwen25mathtechnicalreportmathematical} quality assessment. This process ensures final deliverables meet the same quality standards as the original English data and yields a sizable, high-quality Chinese mathematical dataset. Additionally, we perform full-scale deduplication to eliminate redundancy, guaranteeing dataset uniqueness and validity.

\paragraph{Data Quality Assessment}
To further enhance dataset quality, we establish a robust evaluation framework. Using classifier models \citep{finemath_classifier} and LLM-based \citep{yang2024qwen25mathtechnicalreportmathematical} evaluation methods, we conduct a comprehensive assessment to categorize the dataset into distinct quality groups. This enables strategic removal of low-quality samples and prioritizes the use of high-quality data. Throughout the training pipeline, we strictly maintain these quality standards.
\paragraph{Data Mixing Experiments}
Our experiments are grounded in JT-1.5B-base, a small-scale pretrained LLM homologous to JT-Math-8B-Base. We conduct ablation studies to determine optimal configurations, including mixing ratios between mathematical and other domain data, distribution strategies for mathematical subfields, and training parameters.

\subsubsection{Data Ingredients}\label{subsec:pretrain_valid}

To enhance the general mathematical understanding, reasoning , and long-context learning abilities of the JT-Math Model, we construct a continual pre-training corpus of 210B tokens. This dataset is rigorously filtered and integrated, compiling high-quality content from multiple sources that is heterogeneous and rich in types, including 100B tokens of mathematical concept explanations, computational abilities, and factual mathematical knowledge along with related papers, as well as 110B tokens of mathematical understanding and reasoning, including Chain-of-Thought (CoT), Tool-integrated Reasoning (TIR). In other domains, we primarily select code data covering various programming paradigms and application scenarios based on language model filtering, and add general data from multiple sources, including high-quality web scraper content, open-source books, and specialized texts and industry application documents in the STEM (e.g., Science, Technology, Engineering).

\paragraph{High-Quality Math Data}
\begin{figure}
        \centering
        \includegraphics[width=1\linewidth]{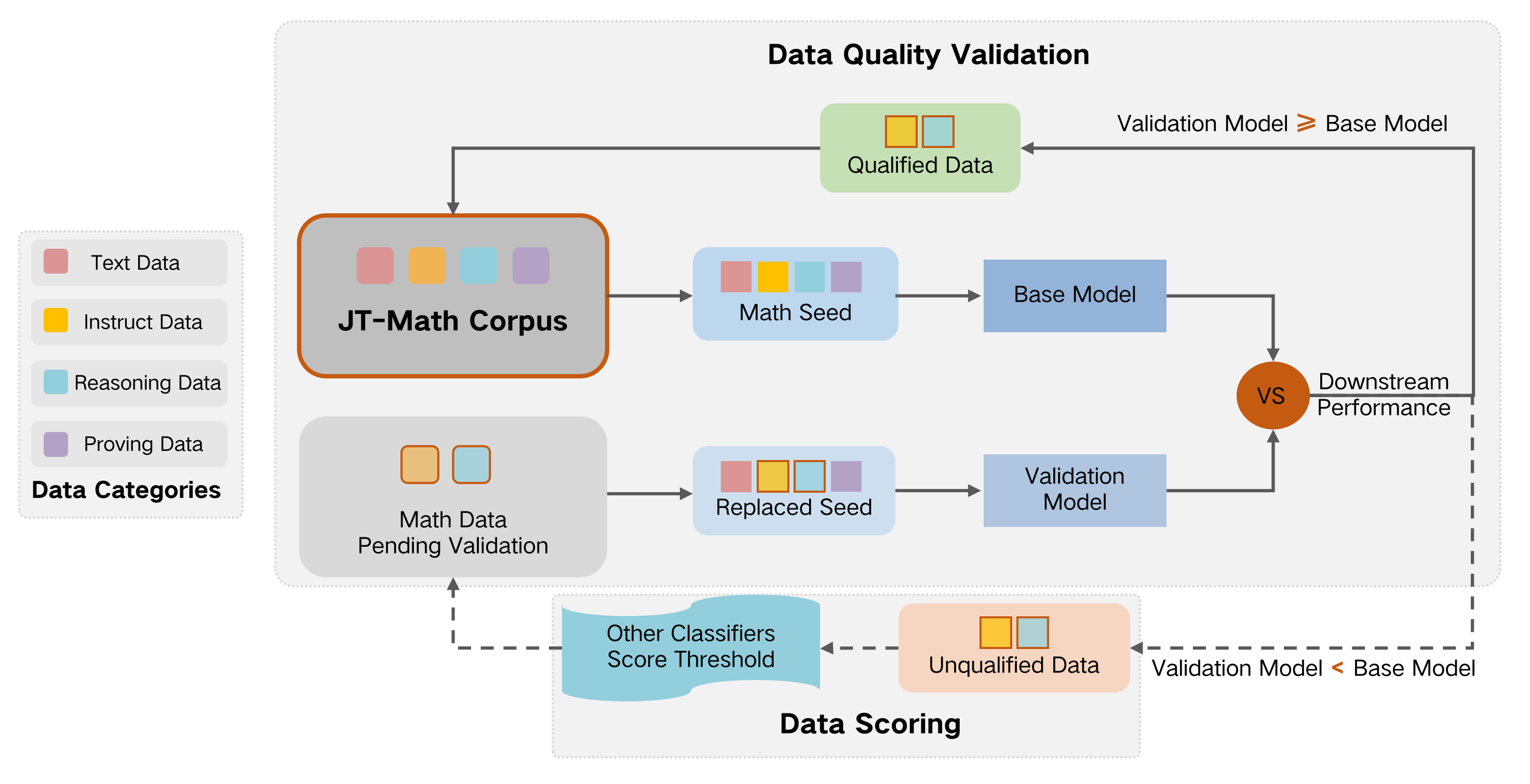}
        \caption{High-Quality Math Data Assessment}
        \label{fig:High-Quality Math Data Evaluation}
    \end{figure}
To improve the quality of data for continual pre-training, we design and implement a high-quality math data assessment process, as shown in Figure ~\ref{fig:High-Quality Math Data Evaluation}. Beyond basic data scoring, this assessment process includes a data quality validation stage. This stage involves replacing data of the same type, training small-scale models, and comparing the downstream performance between the base model and the validation model. This directly measures the data's impact on downstream performance, which in turn guides iterative data optimization.

Initially, we categorize the mathematical data along key influential dimensions to avoid interference from distribution differences. In addition to language type, we further categorize mathematical data into the following categories: text data, instruction data, reasoning data, and proving data. The proportion and total volume of data for each dimension are fixed in subsequent experiments.

To establish a reliable validation baseline, we build high-quality math seed data and train a base model. Using the finemath-classifier \citep{finemath_classifier}, we score the quality of our existing mathematical data and select the highest scoring data 20\% to create the initial JT-Math Corpus. The math seed data is extracted from it via random sampling, with its distribution aligned with the predefined dimensional proportions. Using the math seed data, we perform small-scale continual pre-training on the JT-1.5B-base model to obtain the base model. After a certain quantity is added to the JT-Math Corpus, the seed data and the base model are updated to achieve iterative enhancements in baseline quality and diversity.

For the math data pending validation, we similarly categorize it according to the aforementioned dimensions. We randomly sample an equivalent amount of data of the same type as the seed dataset. This sampled data then replaces the data in the seed dataset (if insufficient, the original data are retained to maintain the total volume). Using the replaced seed, we train a validation model on the same JT-1.5B-base model and with identical hyperparameters (e.g., learning rate, batch-size). This allows for an ablation comparison with the base model.

We then compare the downstream performance differences of the validation model with the base model on the mathematical benchmark. If the validation model outperforms or equals the base model in key metrics, the data are deemed qualified and incorporated into the JT-Math Corpus. Otherwise, we support dynamically adjusting the data scoring strategy, which could include changing classifiers (e.g., open-source LLM scoring \citep{yang2024qwen25mathtechnicalreportmathematical}, finemath-classifier\citep{finemath_classifier}, AutoDS\citep{2024Autonomous}) or raising the classifier's scoring threshold for stricter filtering of the data. To maximize the extraction of potential high-quality data, multiple rounds of iterative validation can be performed. If data quality remains unqualified after more than three rounds, that portion of the data is discarded. Through this dynamic data quality evaluation process, we ultimately obtain 210B tokens of the high-quality JT-Math Corpus.

\paragraph{Long-Context Data}
For math-oriented LLMs, the ability to learn and understand long-context information is crucial for solving complex problems. In the final stage of continued pretraining, we enhance long-context training by extending the sequence length from the original 8,192 tokens to 32,768 tokens. The long context mathematical data primarily consists of LLM-synthesized content and LLM-based selected data. For existing data with sequence lengths exceeding 16,384 tokens, we ensur that the long-context samples were sufficiently challenging. To achieve this, we generate multiple responses for each question using LLMs \citep{yang2024qwen25mathtechnicalreportmathematical} of varying sizes and selected data with a pass rate between $[0.1, 0.9]$ as training samples.

\subsection{Pre-training Stage}

To stabilize the training process and improve model performance, the pre-training phase of the \ourmodel includes three stages:
\paragraph{General Math Knowledge}
During the stage of learning general mathematical knowledge, we initialize the JT-Coder-8B-Base \footnote{JT-Coder-8B-Base employs a from-scratch training approach with Qwen2.5-compatible tokenizer and architecture.}, which improved the model reasoning ability through code training, and then train it on 260B tokens. The distribution of the data is that 70\% is from the JT-Math Corpus, 30\% from AlgebraicStack, code and general knowledge. The sequence length is set to 8192. The learning rate reaches its peak after 2000 warm-up steps, and then uses a constant of $3 \times 10^{-4}$ for the training process.
\paragraph{Reasoning and Thinking}
In order to improve the reasoning and deep thinking abilities of JT-Math, we have utilized more instruction data that has shown significant improvements in downstream performance under quality assesment, as well as Long CoT data with a sequence length of 8192 tokens. We also accelerate the learning rate decay during this stage from $3 \times 10^{-4}$ to $3 \times 10^{-5}$ using cosine scheduling.
\paragraph{Long Context Extension}
In the long context extension stage, we use more long reasoning math and general data, and the long context corpus includes 50\% of text over 8192 tokens in length. The sequence length increases to 32,768 tokens (RoPE base = 500,000) with a learning rate of $3 \times 10^{-5}$.

\section{Post-training: Supervised Fine-Tuning}
\begin{figure}[h]
    \centering
    \includegraphics[height=4cm]{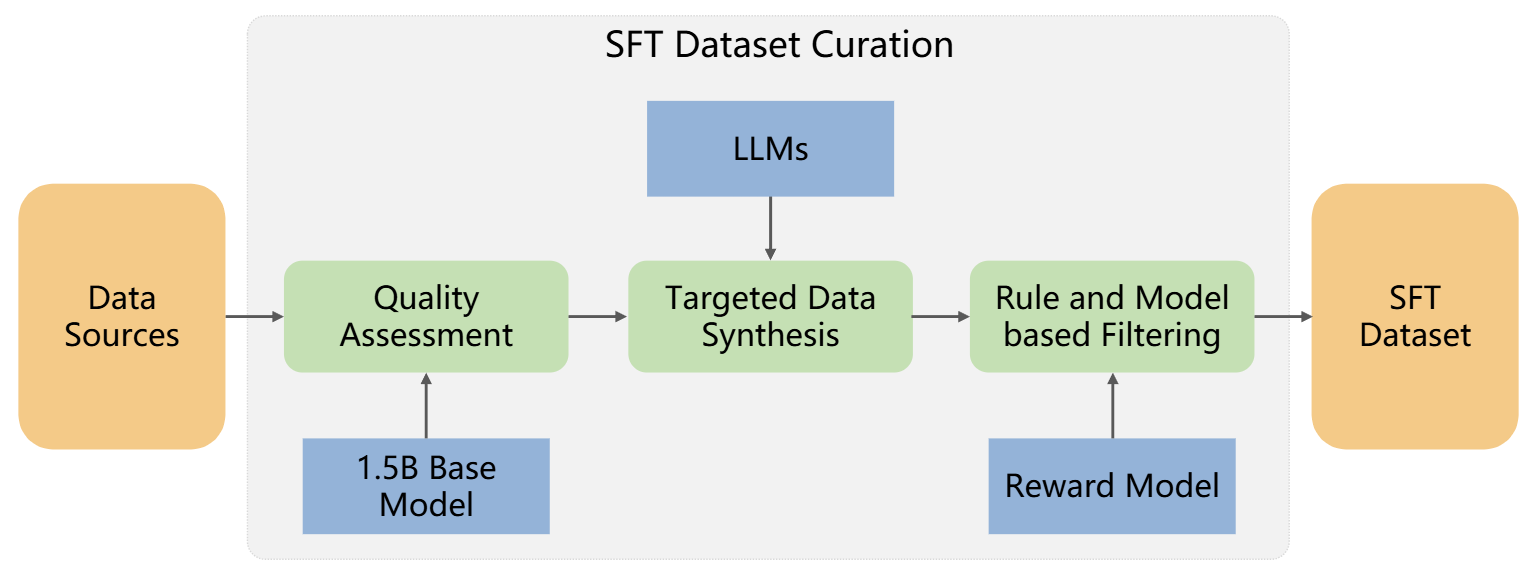}
        \captionsetup{justification=justified, singlelinecheck=false}
    \caption{Our SFT data curation pipeline consists of three key stages: (1) data quality assessment for collected sources using a 1.5B parameter base model; (2) targeted data synthesis to address specific gaps in our collected data; and (3) data filtering based on rule filters and reward model scores.}
    \label{fig:sft_flowchart}
\end{figure}

Our post-training pipeline begins with Supervised Fine-Tuning (SFT), designed to create two specialized models: an \textit{Instruct Model} for clear, concise, and mathematically capable responses, and a \textit{Thinking Model} for tackling complex problems that benefit from extended computational inference. To achieve this, we construct two corresponding SFT datasets: a Short Chain-of-Thought (Short CoT) dataset and a Long Chain-of-Thought (Long CoT) dataset.

As shown in Figure~\ref{fig:sft_flowchart}, We establish a unified data curation pipeline that encompasses three key stages: data quality assessment, targeted data synthesis, rule-based and model-based filtering. Our methodology prioritizes data quality over quantity, starting from a broad collection of open-source datasets and culminating in a highly refined set of several hundred thousand high-quality samples.

\subsection{Data Sources and Quality Assessment}
We begin by aggregating a diverse range of prominent open-source datasets. For Short CoT, this includes NuminaMath \citep{numina_math_datasets}, JiuZhang \citep{NEURIPS2024_0356216f}, ScaleQuest \citep{ding2024unleashing}, and AceMath \citep{liu2024acemath}. For Long CoT, we collect datasets such as OpenThoughts3 \citep{guha2025openthoughtsdatarecipesreasoning}, AM-Thinking-v1 \citep{ji2025amthinkingv1advancingfrontierreasoning}, OpenMathReasoning \citep{moshkov2025aimo2}, Mixture-of-Thoughts \citep{openr1}, and DeepMath \citep{he2025deepmath}.

To identify the highest-quality sources, we conduct a series of validation experiments using a 1.5-billion-parameter base model, which is developed by applying our full pre-training pipeline to the JT-1.5B-Base, as described in Section \ref{sec:pretrain}. We perform SFT on subsets of each dataset, maintaining consistent sample sizes and hyperparameters within each category (Short CoT and Long CoT) to ensure fair comparisons. The quality of each dataset is estimated by measuring the average performance across multiple test benchmarks at several final training checkpoints. Based on these results, we select the top-performing datasets as the foundation for our subsequent data processing and synthesis efforts.

\subsection{Data Synthesis}


Preliminary model evaluations reveal a performance gap on challenging benchmarks like AIME and OlympiadBench when compared to state-of-the-art models. To address this deficiency, we synthesize additional data for difficult problems using two methods. First, we generate new solutions using a more capable model. Second, we leverage an LLM to rewrite and condense the responses generated by DeepSeek-R1 \citep{deepseekai2025deepseekr1incentivizingreasoningcapability}, thereby creating a larger pool of concise, high-quality Short CoT examples for complex problems.

\subsection{Rule-Based Filtering}
We apply a series of rule-based filters to enhance dataset quality. We implement a strict answer-based filtering protocol for all newly synthesized data. For problems lacking a canonical ground truth, we first parse the answer from the original response to serve as a reference. We then extract the answer from our synthesized response and verify its correctness using the math\_verify \citep{Kydlicek_Math-Verify_Math_Verification} library. Samples that fail to yield a parsable answer or whose answer does not match the reference are discarded. 


\subsection{Model-Based Filtering}
In line with findings that data quality is more critical than quantity \citep{zhou2023limaalignment}, and observing that SFT performance tends to saturate, we implement an aggressive model-based filtering strategy. This allows us to retain a small, high-quality subset of the data that achieves performance comparable or superior to the full dataset.

For this task, we employ the Qwen2.5-Math-RM-72B \citep{yang2024qwen25mathtechnicalreportmathematical} reward model to score our Short CoT data. 
Through experimentation, we establish a filtering threshold at the 0.9 quantile, retaining only the top 10\% of samples. 
However, we observe a negative correlation between reward scores and response length. While harder questions typically have longer solutions, a uniform threshold for all samples could inadvertently remove more challenging problems. 
To prevent this, We first group the data based on response token count, using an interval of 128 tokens, then we apply the 0.9 quantile filter independently within each group. 
Ultimately, this strategy successfully reduces the dataset size by 90\% while simultaneously improving performance. 

Through this comprehensive pipeline, we curate high-quality SFT datasets for both Short CoT and Long CoT, each containing several hundred thousand samples.

\subsection{Training Recipe}
The training configurations are tailored for the Instruct and Reasoning models to suit their distinct data types.

\paragraph{Instruct Model}
For the Instruct Model, trained on Short CoT data, we use a learning rate of $3 \times 10^{-6}$, a batch size of 128, and a cosine learning rate scheduler. We utilize the AdamW optimizer with no warmup and set the context length to 8,192 tokens.

\paragraph{Thinking Model}
For the Thinking Model, trained on Long CoT data, we significantly increase the learning rate to $8 \times 10^{-5}$ and the context length to 32,768 tokens, while other hyperparameters remain the same. The higher learning rate is crucial for effectively shifting the model's behavior from its pre-trained state to the complex, step-by-step patterns inherent in the Long CoT data \citep{ji2025amthinkingv1advancingfrontierreasoning}. Our experiments confirm that while a lower learning rate could still elicit similar output patterns, it results in substantially lower accuracy.

\section{POST-TRAINING: Reinforcement Learning}

\begin{figure}[h]
    \centering
    \includegraphics[height=5.27cm]{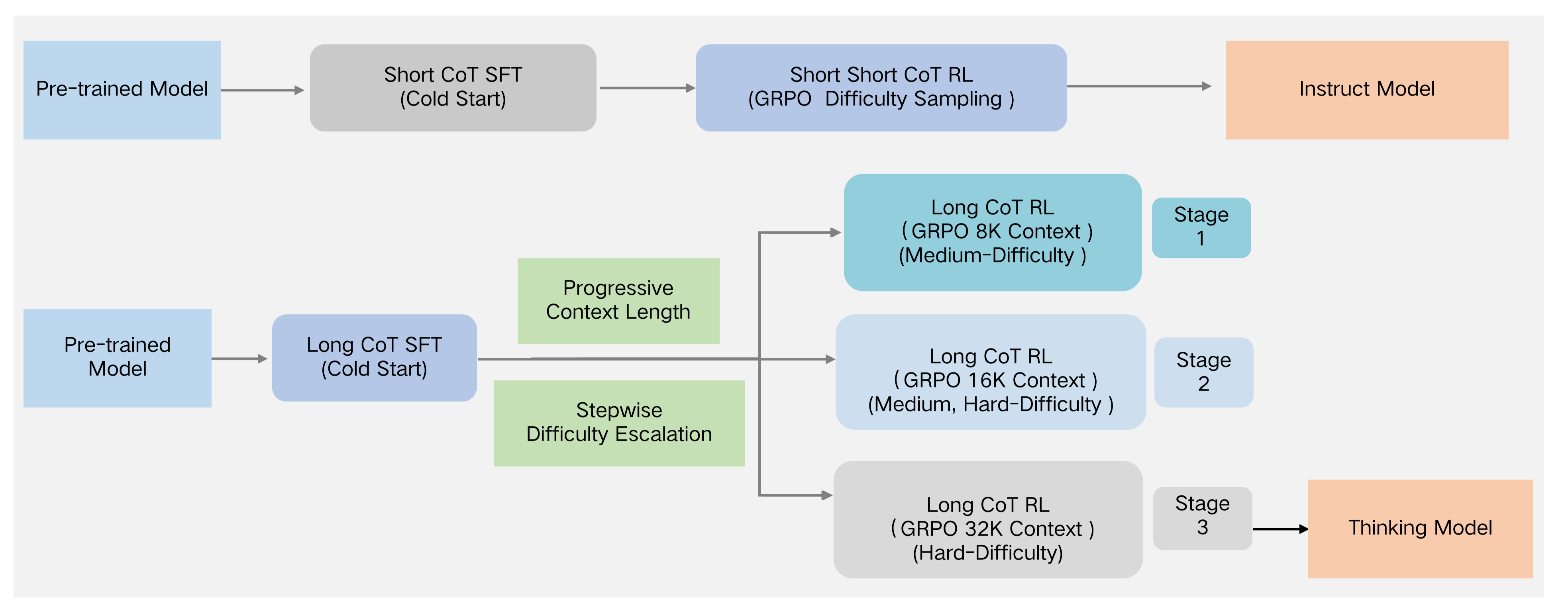}
        \captionsetup{justification=justified, singlelinecheck=false}
    \caption{Our post-training pipeline features two paths:(1) The Instruct Model (Short CoT SFT + RL) provides concise, instruction-following reasoning for moderately difficult problems.
(2) The Thinking Model (Long CoT SFT + multi-stage RL) offers enhanced deep, step-by-step problem-solving in extended contexts.}
    \label{fig:rl_flowchart}
\end{figure}

For reasoning-focused LLMs, SFT often shows limited generalization and reasoning performance, even with high-quality training data. Motivated by these limitations, we extend our investigation to RL as a post-training strategy. Building upon the SFT baseline, we conduct RL experiments aimed at improving the model’s ability to tackle complex mathematical problems and enhance overall performance. Our study evaluates the effectiveness of RL compared to  SFT and explores  training approaches.

\subsection{RL for Instruct Model}

Starting from an SFT-based instruct model, we conduct RL training on short Chain-of-Thought (CoT) tasks to assess the model’s performance under strict output length constraints. This allows us to evaluate its reasoning capability on math problems of moderate and lower difficulty.

\subsubsection{Validation of RL Efficacy}
\begin{figure}[h]
    \centering
    \includegraphics[width=0.6\linewidth]{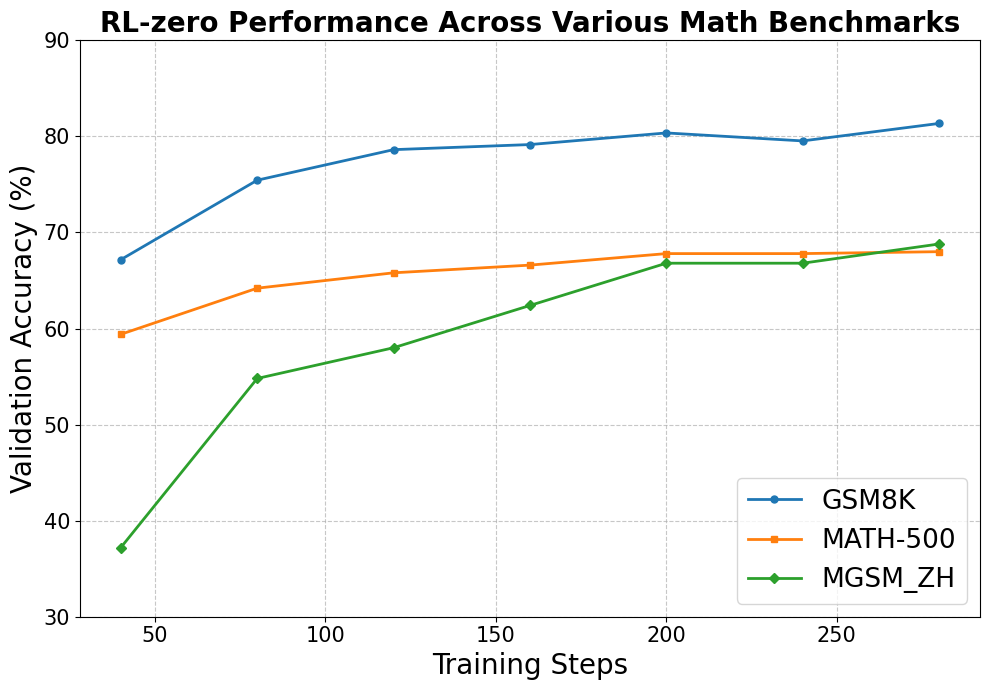}
    \caption{RL-ZERO Performance of CM-1.5B  Base Model during Training on AM-Difficulty-1}
    \label{fig:rl_zero_1.5b}
\end{figure}

We validate RL's effectiveness by directly applying it to the base model (RL-ZERO, in Figure~\ref{fig:rl_zero_1.5b}), observing clear improvements in model performance capabilities. Considering that SFT models offer more consistent behavior in instruction following and output formatting, we adopt  an  SFT + RL training approach that effectively leverages both methods: SFT guides the model on what to say, while RL cultivates its ability to think more deeply.

\subsubsection{Initial  Model Selection}  
The choice of the initial model for RL training has a significant impact on training efficiency. It is initially hypothesized that a highly accurate, near-optimal SFT model might suffer from low policy entropy, thereby limiting the exploration capability of the RL algorithm. To test this hypothesis, we evaluate  multiple SFT checkpoints from different training stages using a consistent RL training protocol. Surprisingly, the results show little variation in performance across different initialization points. This suggests that the best-performing SFT models are are more suitable for further training through RL.

\subsubsection{Data Pipeline}  
Our data strategy involved several key components:  
\begin{itemize}
    \item \textbf{Data Collection:} We select \textbf{DeepMath} \citep{he2025deepmath} and \textbf{AM-Math-Difficulty-RL} \citep{ji2025difficulty} from publicly available datasets after preliminary evaluation.
    \item \textbf{Difficulty Assessment and Dataset Filtering:} To enhance data efficiency, we employ an SFT model to estimate problem difficulty through multiple inferences (N=16) per query. Queries with a correctness rate of exactly 0 or 1 are filtered out to exclude trivial or overly difficult samples.
        \item \textbf{Progressive Context Length Training (8K → 16K → 32K):}
We apply curriculum learning by sequentially training the model at increasing context lengths — starting from 8K tokens, extending to 16K, and ultimately reaching 32K. This gradual increase helps the model adapt to longer reasoning chains while maintaining stability and performance.
    \item \textbf{Multilingual Capabilities:} We introduce  Chinese math data during SFT to build bilingual instruction-following abilities, followed by mixed-language RL training using translated English queries.
\end{itemize}

\subsubsection{Methodological Enhancements}  
To improve the effectiveness and stability of the RL process, we systematically explore and refined several key aspects of the training methodology. We focus on adapting the GRPO algorithm to better suit the mathematical reasoning task and prevent common failure modes such as entropy collapse and suboptimal convergence.

\paragraph{GRPO Improvements}  
We implement and evaluated several modifications to the GRPO algorithm, inspired by recent advances in policy optimization:
\begin{itemize}
    \item Strict On-policy Training: We generate fresh rollouts for each training batch to ensure that policy updates were based on current model behavior, improving gradient consistency and reducing bias from outdated samples.
    \item Static Data Sampling with GRPO:  
To improve training stability and efficiency, we combine static difficulty filtered data with the GRPO algorithm. By pre selecting data based on estimated difficulty levels, we reduce noise in the reward signal and focused training on samples that offered meaningful learning signals.  
    \item Tuning Kullback–Leibler (KL) Divergence Regularization: The KL divergence penalty is reduced to allow the model greater freedom in exploring new solution strategies, helping it escape local optima and discover more effective reasoning paths.
    \item Hyperparameter Explorations: A systematic search is conducted over critical hyperparameters, including response length, rollout count, learning rate, and clip ratio, with the goal of balancing exploration and exploitation while preventing entropy collapse and training instability.
\end{itemize}


\paragraph{Training Recipe}  
We adopt a binary reward function based on symbolic, character-level, and math-verify validations, assigning +1 for correct answers and -1 for incorrect ones.
For Short CoT RL with maximum context length of 8192 tokens, we employ strictly on-policy GRPO with a KL coefficient of $1 \times 10^{-3}$, a learning rate of $4 \times 10^{-6}$, 16 rollouts per query, and a batch size of 256. Prior to training, we perform 16 rollouts per query to estimate solution accuracy and filtered out queries with an accuracy of exactly 0 or 1, representing overly easy or infeasible problems. A sampling temperature of 1.2 is used during generation to encourage diverse generation paths.

\subsection{"Slow Thinking" RL Training }  
Given the limitations of short reasoning chains in handling complex problems, we proceed with RL training on Long CoT generations. This approach provides the model with more space to explore and refine its reasoning steps, enabling deeper problem analysis. We gradually increase the context length from 8K to 16K and finally 32K tokens during training.

\subsubsection{Model and Data}
Building on a strong SFT foundation, we first enhance our model's long-form reasoning through Long CoT training within a 16K context window. This equips the model with the ability to generate step-by-step solutions. We then transition to RL, which encourages deeper, more exploratory thinking, especially crucial for tackling complex problems. Our RL training leverages high-quality datasets like DeepMath \citep{he2025deepmath}, AM-Math-Difficulty-RL \citep{ji2025difficulty}, and AceReason-Nemotron \cite{chen2025acereason},  providing verified ground-truth solutions for reliable supervision.

\subsubsection{Training Methods}
Our training methodology is designed to carefully balance training efficiency with final performance. We begin with \textbf{Static Data Sampling with GRPO } as a time-efficient baseline. And we employ a \textbf{multi-stage RL process to progressively expand the context window}.

This process begins after the initial 16K SFT model has been obtained. For reasons of both efficiency and performance, the RL training itself commences at a shorter, more manageable context window of 8K. This allows the model to quickly learn the reward-driven behaviors with faster iterations and lower computational overhead. Once the model achieves a stable level of performance at 8K, we expand the training context to 16K, and subsequently to a target of 32K, using the previously trained checkpoint as a starting point for each new stage.

This gradual increase in context length functions as a practical curriculum. It allows the model to solidify its reasoning capabilities at each stage before tackling the increased complexity and longer dependencies inherent in larger context problems. This method is significantly more resource-efficient than starting RL directly at the maximum context length and leads to more stable training and robust performance. 

For the Thinking Model trained on Long CoT data, we refine the training set by performing 16 rollouts per query to estimate accuracy and filtered out queries with scores of exactly  1. We then applied strictly on-policy GRPO with a KL coefficient of $1 \times 10^{-3}$, learning rate $4 \times 10^{-6}$, 16 rollouts per query, batch size 32. We use a sampling temperature of 1.2 to encourage diverse reasoning paths.

\section{Decontamination}
To ensure the validity of our results by preventing test set contamination, we performed a comprehensive decontamination of all data utilized in both pre-training and post-training stages. For this purpose, we employed the established 10-gram filtering methodology. This approach identifies and eliminates any training instance containing a 10-gram sequence that is also present in key benchmark datasets.

\section{Results}

We conducted an extensive series of evaluations to investigate the performance of the following models: 1) \ourmodel-Base, the pretrained foundation model; 2) \ourmodel-Instruct, an instruction-tuned variant optimized for fast inference on simpler tasks; 3) \ourmodel-Thinking,  a specialized model using reinforcement learning for complex problems that require longer inference. In the sections that follow, we present detailed results and benchmark their performance against previous state-of-the-art open-source LLMs.

\subsection{Benchmarks}

To evaluate our \ourmodel models, we employ a suite of benchmarks spanning a wide spectrum of mathematical difficulty, from elementary arithmetic to international olympiad-level challenges.

\textbf{GSM8K  \citep{cobbe2021training}.}
GSM8K is a dataset of linguistically diverse, multi-step word problems from grade school mathematics. It primarily evaluates a model's foundational arithmetic and step-by-step reasoning. Strong performance on GSM8K indicates reliable logical inference for problems requiring a sequence of 2-to-8 solution steps.

\textbf{Math-500. \citep{lightman2023let}}
The Math (500-problem) set elevates the evaluation to formal proofs and symbolic reasoning. This collection of intermediate-difficulty problems requires models to generate detailed, step-by-step solutions.

\textbf{AIME 2024. \citep{MAA:AIME2024}}
The AIME (American Invitational Mathematics Examination) is a U.S. high school mathematics competition with a difficulty between the AMC and USAMO. We use the AIME 2024  to assess performance on competition-level mathematics that requires deep insight and sophisticated skills. These problems demand creative and advanced reasoning, as they often lack standard solution templates.

\textbf{AIME 2025. \citep{MAA:AIME2025}}
Building on the challenge of AIME 2024, the AIME 2025 dataset further assesses performance on competition-level mathematics requiring deep insight and sophisticated skills. These problems demand creative and advanced reasoning, as they often lack standard solution templates, pushing the boundaries of a model's ability to handle novel and complex mathematical challenges.

\textbf{Olympiad Bench. \citep{he2024olympiadbench}}
The Olympiad Bench is a curated collection of problems from premier competitions like the International Mathematical Olympiad (IMO) and the Putnam Competition. These problems represent the pinnacle of pre-university mathematics and are designed to be exceptionally difficult. Success requires synthesizing knowledge across fields like number theory and geometry and generating novel proof strategies. Performance on this benchmark measures a model's potential to approach expert-level mathematical reasoning.

\textbf{AMC 23. \citep{MAA:AMC}}
 The AMC (American Mathematics Competitions) are a series of examinations that aim to identify and foster mathematical talent in high school students. Specifically, AMC 23 (referring to the 2023 competition) problems serve as a benchmark to assess a model's foundational mathematical problem-solving skills, including algebra, geometry, number theory, and counting. These problems range in difficulty and are designed to test both conceptual understanding and computational accuracy under timed conditions.

\textbf{CNMO 2024. \citep{CNMO2024}}
 The CNMO (Chinese National Mathematics Olympiad) is a highly prestigious mathematics competition for high school students in China. Problems from the CNMO 2024 are used to evaluate a model's ability to solve complex, non-routine mathematical problems that often require innovative thinking, deep theoretical understanding, and multi-step reasoning, reflecting the high standards of mathematical education and competition in China.

\textbf{CMath. \citep{Wei2023CMATH}}
CMath is a benchmark of mathematical problems sourced from Chinese school competitions and assignments. It assesses mathematical computation and reasoning while specifically testing the model's ability to interpret problems posed within complex Chinese linguistic contexts.


\subsection{Baselines}
We compare our \ourmodel with the following baselines.

\textbf{DeepSeek-Math} \citep{shao2024deepseekmathpushinglimitsmathematical} is a family of 7B language models specialized in mathematical reasoning, built upon the DeepSeek-Coder-Base model. These models undergo continual pre-training with a 120B-token corpus of high-quality mathematical web data sourced from Common Crawl. The series includes an instruction-tuned variant, DeepSeek-Math-Instruct, trained on problems with Chain-of-Thought and Program-of-Thought solutions, and an RL-tuned model, DeepSeek-Math-RL. The latter is enhanced using Group Relative Policy Optimization (GRPO), a memory-efficient reinforcement learning algorithm that significantly boosts performance on mathematical benchmarks like MATH, achieving results competitive with leading closed-source models.

\textbf{Qwen2.5-Math} \citep{yang2024qwen25mathtechnicalreportmathematical} is an enhanced mathematical reasoning model based on the Qwen2-Math series, featuring foundational models with 1.5B/7B/72B parameters and instruction-tuned variants. It incorporates synthetic math data from Qwen2-Math during pretraining and employs reward models to improve reasoning alignment. The model demonstrates significant improvements in mathematical problem-solving, programming tasks, and instruction-following, supported by multilingual capabilities covering 29+ languages.


\textbf{DeepSeek-R1-Distill-Qwen} \citep{deepseekai2025deepseekr1incentivizingreasoningcapability} is a series of distilled large language models based on the Qwen2.5 series, utilizing reasoning data generated by DeepSeek-R1. Ranging from 1.5B to 32B parameters, these models are optimized for math, code, and reasoning tasks, demonstrating exceptional efficiency and accuracy, often outperforming larger models, and providing a distilled yet powerful baseline for Math LLM evaluation.

\textbf{GPT-4o} \citep{openai2024gpt4ocard} is a state-of-the-art large language model developed by OpenAI, renowned for its advanced capabilities in natural language understanding, generation, and multimodal processing, including text and image inputs. GPT-4o serves as a leading general-purpose baseline for assessing performance across a wide range of tasks, including mathematical reasoning. We evaluate the GPT-4o-1120-128k version.

\textbf{OpenAI o1-mini} \citep{openai2024openaio1card} is a smaller, cost-effective model from OpenAI's o1 series, specifically designed for technical problem-solving in areas like coding, mathematics, and logical reasoning. It maintains strong reasoning abilities due to training with large-scale reinforcement learning for Chain-of-Thought. O1-mini prioritizes efficiency and speed, providing quick responses for complex, specialized tasks, and performs well on benchmarks such as AIME and HumanEval.

\subsection{Evaluation of Base Models}

We evaluate \ourmodel using few-shot CoT prompting on established mathematical benchmarks: GSM8K, MATH, and CMATH. 
In addition to the baseline math models mentioned above (DeepSeek-Math-Base-7B and Qwen2.5-Math-7B), our comprehensive comparison further include general-purpose language models (Qwen2.5-Base-32B \citep{qwen2025qwen25technicalreport} and Llama-3.1-Base-405B \citep{llama3modelcard}) and code models ( DeepSeek-Coder-V2-Lite-Base \citep{deepseekai2024deepseekcoderv2breakingbarrierclosedsource}) to thoroughly benchmark performance. 
The experimental results presented in Table ~\ref{tab:benchmark_performance} demonstrate \ourmodel superior mathematical reasoning abilities, achieving scores of 60.1 on MATH and 90.1 on CMath, with an average score that significantly outperforms all baseline models of similar scale.

\begin{table}[h]
\centering
    \caption{Model Performance Comparison Across Multiple Math Benchmarks.}
    \label{tab:benchmark_performance}
\begin{tabular}{l|cccc}
\hline
  \textbf{Model} & \textbf{\makecell{GSM8K}} & \textbf{\makecell{MATH}}& \textbf{\makecell{CMath\\(zh)}} & \textbf{Average} \\ 
        \midrule
        \multicolumn{5}{c}{\textit{General Model}} \\
        \midrule
        Qwen2.5-Base-32B & \textbf{92.8} & 57.7 & 85.4& 78.6\\
        Llama-3.1-Base-405B & 89.0&  53.8&  77.4& 73.4\\
        \midrule
        \multicolumn{5}{c}{\textit{Specific Model}} \\
        \midrule
        DeepSeek-Math-Base-7B & 64.2 & 36.2 & 71.7 & 57.4\\
        DeepSeek-Coder-V2-Lite-Base & 68.3 & 38.1 & 77.8 & 61.4 \\
        Qwen2.5-Math-7B & 91.6 & 55.4 & 85.0 & 77.3\\ 
        \textbf{\ourmodel-Base}  & 87.5& \textbf{60.1}& \textbf{90.1}& \textbf{79.2}\\
        \bottomrule
    \end{tabular}
\end{table}

\subsection{Evaluation of Instruct Models}

\begin{table}[h]
\centering
\caption{Model Performance Comparison Across Multiple Math Benchmarks.}
\label{tab:instruct}
\resizebox{\textwidth}{!}{%
\begin{tabular}{l|ccccccccc}
\hline
\textbf{Model} & \textbf{\makecell{MATH-500}} & \textbf{\makecell{AIME\\24}} & \textbf{\makecell{AIME\\25}} & \textbf{\makecell{Olympiad\\Bench}} & \textbf{\makecell{AMC\\23}} & \textbf{\makecell{CNMO\\24}} & \textbf{\makecell{CMath}} & \textbf{Average} \\ \hline
DeepSeek-Math-7B-Instruct    & 40.60 & 0.00 & 0.00 & 11.31 & 14.38 & 0.00 & 82.06 & 21.19 \\
Qwen2.5-Math-7B-Instruct & 82.40 & 13.33 & 16.67 & 34.66 & 62.50 & 27.78 & \textbf{92.53} & 47.12 \\
GPT-4o & 74.20 & 12.92 & 10.42 & 38.09 & 49.06 & 20.14 & 72.50 & 39.62 \\
\textbf{\ourmodel-Instruct} & \textbf{90.00}  & \textbf{37.50} & \textbf{34.17} & \textbf{47.04} & \textbf{76.56} & \textbf{69.44} & 91.44 & \textbf{63.74}  \\
\hline
\end{tabular}%
}
\end{table}

\textbf{Evaluation Settings.} We use greedy decoding for all instruct models. We set the maximum output length to 8,192 tokens for our \ourmodel-Instruct, 16,384 tokens for GPT-4o, and 4,096 tokens for DeepSeek-Math-7B-Instruct and Qwen2.5-Math-7B-Instruct as their context length is limited to 4,096.

The evaluation results of instruct models are presented in Table \ref{tab:instruct}. The results highlight that \ourmodel-Instruct achieves superior performance across a comprehensive suite of mathematical reasoning benchmarks. Notably, JT-Math-8B-Instruct achieves the highest scores on nearly every individual benchmark, securing top marks of 90.00 on MATH-500, 37.50 on AIME 24, 34.17 on AIME 25, 47.04 on Olympiad Bench, 76.56 on AMC 23, and an impressive 69.44 on the challenging CNMO 24. This consistent dominance culminates in a remarkable average score of 63.74, which is over 16 points higher than the next best performing model, Qwen2.5-Math-7B-Instruct (47.12). This significant margin underscores the model's robust and generalized mathematical capabilities, establishing its superior proficiency in solving a wide spectrum of mathematical problems, from foundational tests to complex, competition-level challenges.

\subsection{Evaluation of Thinking Models}

\begin{table}[h]

\centering

\caption{Model Performance Comparison Across Multiple Math Benchmarks.}

\label{tab:reasoning}

\resizebox{\textwidth}{!}{%

\begin{tabular}{l|ccccccccc}

\hline

\textbf{Model}  & \textbf{\makecell{MATH-500}} & \textbf{\makecell{AIME\\24}} & \textbf{\makecell{AIME\\25}} &\textbf{\makecell{Olympiad\\Bench}}    &\textbf{\makecell{AMC\\23 }} & \textbf{\makecell{CNMO\\24}} &\textbf{\makecell{CMath}}  & \textbf{Average} \\ \hline

DeepSeek-R1-Distill-Qwen-7B  & 92.00 & 54.64 & 41.30 &54.62 &90.62  &65.45  & 88.62  &69.61  \\

o1-mini-128k        & 90.20 & 61.25 & 49.17 &51.07 &91.56 &52.78 &89.80  &69.40  \\

\textbf{\ourmodel-Thinking} &\textbf{93.80} &\textbf{69.17} &\textbf{58.75} &\textbf{59.59} &\textbf{96.25} &\textbf{72.22} &\textbf{93.99} &\textbf{77.68}\\

\hline

\end{tabular}%

}

\end{table}


\textbf{Evaluation Settings.} We use sampling for all reasoning models. We set the maximum generation length to 32,768 tokens and temperature to 0.65 for our JT-Math-8B-reasoning, 32,768 tokens and 0.6 temperature for DeepSeek-R1-Distill-Qwen-7B, and 32,768 tokens for o1-mini-128k. We use the default temperature for o1-mini.

We assess the JT-Math-8B-Thinking model on various challenging mathematical reasoning benchmarks, with average@8 scores summarized in Table \ref{tab:reasoning}. As the table shows, JT-Math-8B-Thinking delivers strong competitive performance, achieving the highest average score across diverse benchmarks. It particularly excels in challenging math competition problems, demonstrating a substantial edge over other models. Furthermore, its robust capabilities extend to formal reasoning tasks and Chinese mathematical benchmarks, where it surpasses other models in its class. These results collectively highlight JT-Math-8B-Thinking's effective performance and impressive problem-solving prowess across multilingual and varied mathematical reasoning scenarios.

\section{Conclusion, Limitation, and Future Work}

In this paper, we introduce \ourmodel, a series of open-source large language models designed for advanced mathematical reasoning. We demonstrate that a systematic, multi-stage optimization framework, encompassing both pre-training and post-training, can significantly enhance a model's mathematical capabilities. Exposed to diverse mathematical knowledge patterns during its pre-training, \ourmodel establishes a strong foundation for its exceptional reasoning potential. In the post-training phase, through supervised fine-tuning and a curriculum reinforcement learning pipeline, the model achieves state-of-the-art performance on several mainstream math benchmarks, surpassing models of comparable and even larger scales.

As the training of \ourmodel primarily focuses on mathematical tasks, the model's capabilities in general natural language understanding and handling broader tasks are comparatively limited. Furthermore, a data-scale gap exists compared to leading models trained on trillions of tokens, which may impose inherent limitations on the model's breadth of knowledge. Our future work will focus on exploring the application of this framework to multi-tool mathematical reasoning and other scientific domains that demand strong logical abilities, aiming to continually enhance its performance and scope.

\bibliography{main_bibliography}
\bibliographystyle{main_bibliography}


\end{document}